\documentclass[journal]{IEEEtran}
\usepackage{array} 
\newcommand{\specialcell}[2][c]{%
	\begin{tabular}[#1]{@{}c@{}}#2\end{tabular}}
\usepackage[hidelinks]{hyperref}
\usepackage{xurl}
\usepackage{subcaption}
\usepackage{cite}
\usepackage{amsmath,amssymb,amsfonts}
\usepackage{algorithm}
\usepackage{algpseudocode}
\usepackage[font=small,labelfont=bf,labelsep=colon]{caption}
\makeatletter
\renewcommand{\fnum@algorithm}{\textbf{Algorithm~\thealgorithm:}}
\makeatother
\usepackage{graphicx}
\usepackage{textcomp}
\usepackage{xcolor}
\usepackage{algorithm}
\usepackage{pgfplots}
\usepackage{epstopdf}
\usepackage[T1]{fontenc}
\usepackage{url}
\pgfplotsset{width=\linewidth,compat=1.9}
\def\BibTeX{{\rm B\kern-.05em{\sc i\kern-.025em b}\kern-.08em
    T\kern-.1667em\lower.7ex\hbox{E}\kern-.125emX}}
\begin{document}

\title{Vision-Aided Online A$^*$ Path Planning for Efficient and Safe Navigation of Service Robots}

\author{Praveen Kumar, and Tushar Sandhan, \textit{Senior Member, IEEE}%
	\thanks{Praveen Kumar and Tushar Sandhan are with the Department of Electrical Engineering, Indian Institute of Technology Kanpur, India (e-mail: praveenk20@iitk.ac.in; sandhan@iitk.ac.in).}%
}

\maketitle

\begin{abstract}
The deployment of autonomous service robots in human-centric environments is hindered by a critical gap in perception and planning. Traditional navigation systems rely on expensive LiDARs that, while geometrically precise, are semantically unaware, they cannot distinguish a important document on an office floor from a harmless piece of litter, treating both as physically traversable. While advanced semantic segmentation exists, no prior work has successfully integrated this visual intelligence into a real-time path planner that is efficient enough for low-cost, embedded hardware. This paper presents a framework to bridge this gap, delivering context-aware navigation on an affordable robotic platform. Our approach centers on a novel, tight integration of a lightweight perception module with an online A* planner. The perception system employs a semantic segmentation model to identify user-defined visual constraints, enabling the robot to navigate based on contextual importance rather than physical size alone. This adaptability allows an operator to define what is critical for a given task, be it sensitive papers in an office or safety lines in a factory, thus resolving the ambiguity of what to avoid. This semantic perception is seamlessly fused with geometric data. The identified visual constraints are projected as non-geometric obstacles onto a global map that is continuously updated from sensor data, enabling robust navigation through both partially known and unknown environments. This process creates a unified map of both physical and semantic objects, upon which the online A* search algorithm computes optimal, collision-free paths that intelligently respect this comprehensive world representation. We validate our framework through extensive experiments in high-fidelity simulations and on a real-world robotic platform. The results demonstrate robust, real-time performance, proving that a cost-effective robot can safely navigate complex environments while respecting critical visual cues invisible to traditional planners. Our work provides a practical and scalable solution for deploying the next generation of truly intelligent service robots. The source code and dataset will be publicly available.
\end{abstract}



\begin{figure}[t!]
	\centering
	\includegraphics[width=1.0\columnwidth, trim={5.0cm 2.3cm 5.0cm 2.0cm}, clip]{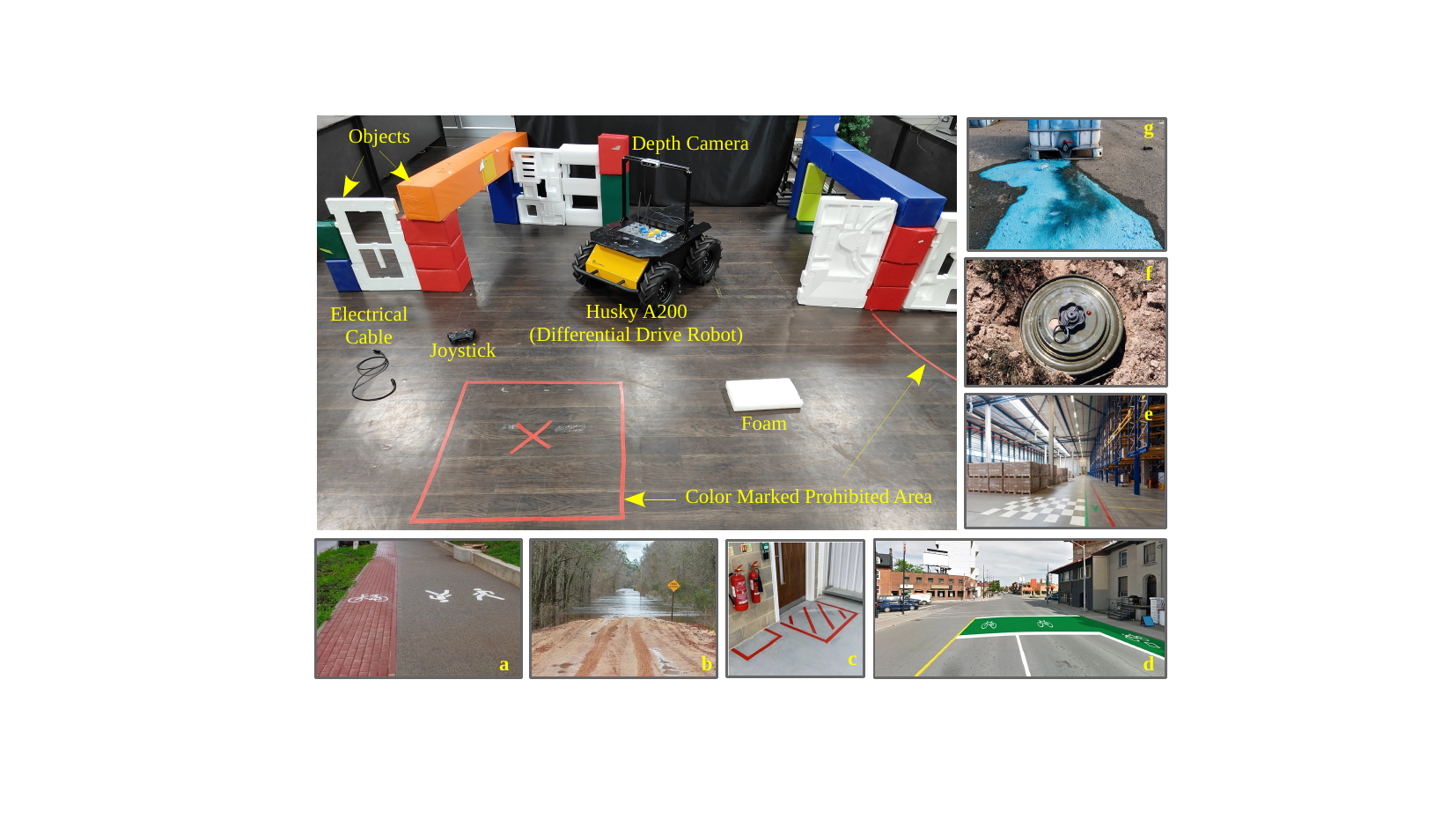}
	\caption{Illustrates the real-world setups and scenarios: Husky A200 wheeled robot equipped with an Intel RealSense D455 depth camera, navigating an indoor environment with avoidable items (e.g., foam), unavoidable items (e.g., electrical cables, joystick), and a color-marked prohibited area. The robot encounters several challenges: interpreting color-coded markings and symbols to identify prohibited areas presented in label (a); distinguishing muddy water at floor level to prevent misclassification as navigable terrain (b); avoiding reserved spaces marked with specific colors (c); recognizing user-defined color-coded ground surfaces indicative of organized zones (d); identifying industry-standard color markings denoting workstations for safety and critical operations (e); detecting potential landmines embedded within the navigation path (f); and recognizing chemical spills present on the surface (g). Each scenario is integrated into a comprehensive knowledge graph, enabling \textbf{our} algorithm to plan safe and user-centric navigation paths.}
	\label{fig:rw}
\end{figure}

\section{Introduction}
Path planning for ground robots in complex, human-centric environments is a highly engaging research topic, attracting significant attention from both academia and industry. Traditionally, 3D LiDAR sensors have been the cornerstone of autonomous navigation, providing precise geometric data for obstacle avoidance. However, these systems suffer from two critical drawbacks for the widespread adoption of service robots: they are prohibitively expensive, and they are fundamentally \textit{semantically blind}. They perceive the world as a collection of points in space but lack the ability to interpret color, texture, or symbolic information. This limitation is a major barrier in real-world scenarios where visual cues are paramount. For example, a LiDAR-based robot cannot distinguish muddy water from a solid floor, interpret a color-coded safety line in a factory, or recognize a chemical spill on the ground.  These non-geometric, semantic hazards are invisible to purely geometric sensors, posing significant risks to the robot and its mission.

The need to incorporate visual intelligence has led to a dominant trend in modern robotics: the use of data-driven, learning-based methods. Recent breakthroughs, such as the DiPPeR~\cite{liu2024dipper} and NoMaD~\cite{sridhar2024nomad} frameworks, demonstrate impressive capabilities in generating smooth, learned trajectories directly from visual data. However, these state-of-the-art approaches present their own set of limitations for practical, low-cost service robotics. Firstly, they are computationally expensive, often benchmarked on powerful GPUs that are not viable for affordable, embedded platforms. Secondly, and more critically, these learned policies are "black boxes" that are not designed to incorporate explicit, arbitrary, and user-defined symbolic rules with 100\% certainty. Enforcing a strict, non-negotiable command like "do not cross this temporary wet floor sign"—a constraint unseen during training—is non-trivial and cannot be guaranteed. This leaves a crucial gap for a framework that can reliably and affordably enforce semantic rules in real-time.

This paper bridges this gap by presenting a novel, hybrid framework that tightly integrates context-aware visual intelligence with a real-time, classical path planner. Our approach operates using only a single, low-cost RGB-D sensor, making it an economically viable alternative to expensive LiDAR systems. We enhance the proven robustness of an online A* planner operating on a dynamic occupancy grid with a lightweight perception module. We leverage the efficient ESANet~\cite{seichter2021efficient} semantic segmentation model to perceive the environment not merely as a collection of obstacles, but as a space rich with user-defined meaning. Crucially, by fusing semantic data from the RGB stream with geometric data from the depth stream, our system can robustly differentiate between real, physical objects (e.g., a teacup) and deceptive visual elements (e.g., a photograph of a teacup), a vital capability for reliable real-world operation.

Our framework is designed to provide adaptable intelligence. It allows an operator to define a set of critical visual constraints specific to a mission—be it sensitive documents in an office, safety zones in a factory, or potential threats in a defense scenario. The perception module identifies these user-defined constraints, which are then projected as non-geometric obstacles onto the robot's occupancy grid map. This process creates a unified world representation containing both physical barriers and semantic rules. The online A* planner then computes an optimal, collision-free path that respects this comprehensive map, ensuring the robot navigates safely and intelligently according to its mission objectives. This hybrid approach combines the mathematical rigor of a classical planner with the perceptual power of a modern deep learning model.

The key contributions of this paper are as follows:
\begin{enumerate}
	\item \textbf{A Novel Hybrid Planning Framework:} We present the first practical integration of a lightweight semantic perception module with a dynamic, grid-based online A* planner, enabling cost-effective robots to navigate based on a unified map of both geometric and non-geometric, user-defined constraints.
	\item \textbf{Demonstration of Real-Time Semantic Awareness on Embedded Hardware:} We prove that robust, context-aware navigation is feasible without relying on expensive, power-intensive GPUs, by pairing the efficient ESANet model with a classic planner on a low-cost robotic platform.
	\item \textbf{An Open-Source Implementation and Dataset:} We will publicly release our framework and a curated dataset of annotated service robotics scenarios to foster further research in accessible, real-world semantic navigation.
\end{enumerate} 
\begin{figure*}[t!]
	\centering
	\includegraphics[width=2.0\columnwidth, trim={0.0cm 5cm 0.5cm 2.5cm}, clip]{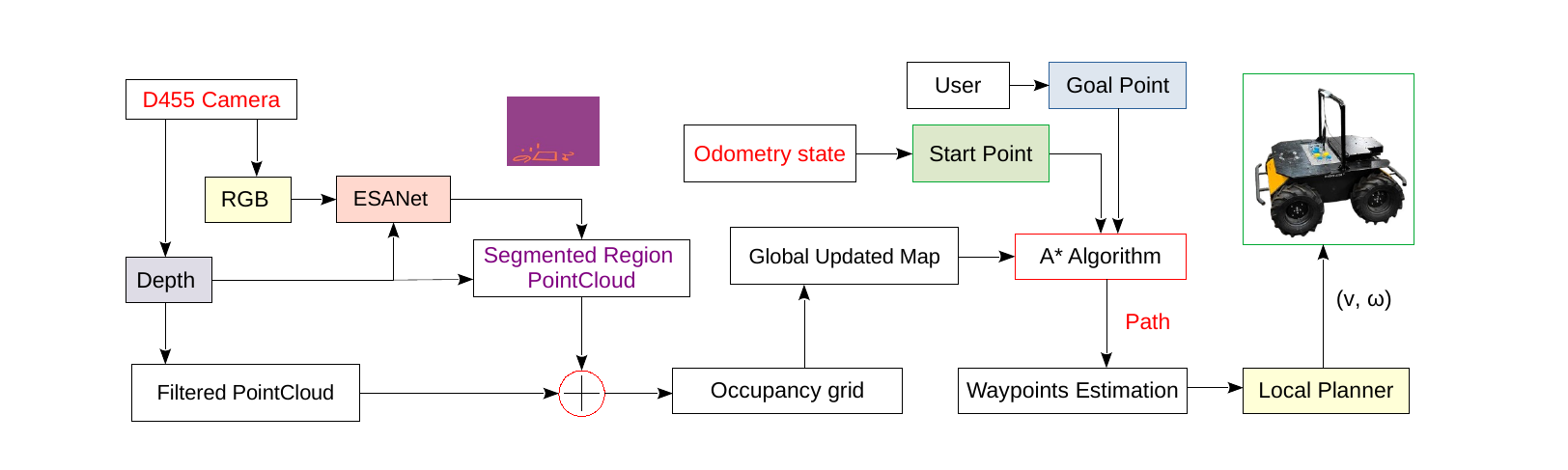}
	\caption{System Architecture Overview: Data from the D455 RGB-D camera is processed in two streams. The RGB and depth streams feed the ESANet semantic segmentation model, whose output is combined with depth data to generate a segmented region point cloud. The depth stream provides a filtered point cloud representing geometric obstacles. Both point clouds are fused into a dynamic occupancy grid, updating the global map. An online A* algorithm uses this map, along with the robot's current state (Odometry) and user-defined goal, to compute a global path. This path is converted into waypoints for a local planner, which generates final velocity commands $(v, \omega)$ for the robot.}
	\label{fig:rednet}
\end{figure*}
    
\section{RELATED WORK}
Path planning for autonomous service robots requires a delicate balance between safety, efficiency, and environmental understanding. The research landscape can be broadly categorized into three key areas relevant to our work: classical geometric path planning, modern learning-based motion planning, and the lightweight perception models that enable environmental intelligence. In this section, we critically review the state-of-the-art in these domains to establish the crucial gap that our proposed framework addresses.

\subsection{Classical and Geometric Path Planning}
Classical path planning has traditionally focused on finding optimal, collision-free paths through environments represented by geometric primitives. These methods are typically divided into search-based and sampling-based approaches.

Search-based planners, such as A*~\cite{hart1968formal} and D* Lite~\cite{koenig2002d}, provide mathematical guarantees of finding an optimal solution. They operate on discretized representations like the occupancy grid—the paradigm adopted in our work—and are valued for their reliability in unknown environments. Sampling-based planners like RRT~\cite{lavalle1998rapidly} efficiently explore high-dimensional free spaces but lack optimality guarantees, while its variant RRT*~\cite{karaman2011sampling} introduces asymptotic optimality through incremental rewiring. An alternative state-of-the-art approach uses visibility graphs, as seen in planners like the FAR Planner~\cite{yang2022far}, which show exceptional speed in sparse environments. However, their performance can degrade in cluttered spaces typical of indoor service robotics.

Despite their algorithmic differences, a fundamental limitation unites all these classical methods: \textbf{they are semantically blind}. They perceive the world as a binary collection of free space versus occupied space. An A* planner on an occupancy grid is just as incapable of understanding a red "keep-out" zone as a visibility graph planner is. This semantic blindness severely limits their utility for service robots that must operate intelligently in complex, human-populated environments.

\subsection{State-of-the-Art in Learning-Based Motion Planning}
To overcome the rigid nature of classical planners, a dominant trend in modern robotics is the use of data-driven, learning-based methods that generate smooth and dynamically feasible trajectories.

Recent breakthroughs in this area, such as the DiPPeR~\cite{liu2024dipper} framework, leverage diffusion models to learn a direct mapping from a 2D map image to a global trajectory. This approach is powerful and, as the authors show, can be significantly faster than traditional search-based methods. The follow-up work, DiPPeST~\cite{stamatopoulou2024dippest}, extends this by using the generated path for real-time local refinement from camera images. Similarly, NoMaD~\cite{sridhar2024nomad} presents a unified diffusion policy that can handle both goal-directed navigation and exploration in unseen environments.

However, despite their power, these learning-based systems have fundamental limitations for our target application. First, they are computationally intensive. DiPPeR, for example, was benchmarked on a powerful NVIDIA RTX 3090 GPU, a class of hardware not viable on most low-cost service robot platforms. Second, and more critically, they are designed to learn from geometric or visual features, not explicit symbolic rules. Their policies are trained on large datasets of successful trajectories. Enforcing a strict, non-negotiable, and arbitrary rule like "do not cross this temporary wet floor sign"—a constraint that was not present in the training data—is non-trivial and cannot be guaranteed. Their strength lies in generalizing learned behaviors from visual patterns, not in adhering to explicit, user-defined semantic constraints.

\subsection{Lightweight Semantic Segmentation for Robotics}
The performance of our framework is contingent on its perception model. While large-scale models like SegFormer~\cite{xie2021segformer} offer top-tier accuracy, their computational demands are prohibitive for real-time robotics on embedded hardware. The challenge is to find an architecture that maximizes accuracy while minimizing latency.

We analyzed several prominent lightweight architectures. The classic U-Net~\cite{ronneberger2015u} is a popular choice but is often too slow without significant pruning. More recent models like MIPANet~\cite{zhang2024mipanet} show strong performance but can have a higher memory overhead. Our selection, ESANet~\cite{seichter2021efficient}, was designed specifically for real-time performance. It employs efficient, factorized convolutions to capture rich context with minimal computational cost. As argued in Table 1, ESANet provides a superior trade-off between inference speed and accuracy on a representative embedded platform, making it the ideal choice for our framework.

\begin{table}[htbp]
	\centering
	\renewcommand{\arraystretch}{1.15} 
	\setlength{\tabcolsep}{5pt}         
	\footnotesize                      
	
	\begin{tabular}{@{}l l c c c@{}}
		\hline
		\textbf{Method} & \textbf{Backbone} & \textbf{NYUv2} & \specialcell{\textbf{SUN}\\\textbf{RGB-D}} & \textbf{FPS} \\
		\hline
		FuseNet~\cite{hazirbas2016fusenet}                 & 2×VGG16        & —      & 37.3    & $\dagger$ \\
		RedNet~\cite{jiang2018rednet}                  & 2×R34          & —      & 46.8    & 26.0 \\
		SSMA                    & 2×mod.R50      & —      & 44.4    & 12.4 \\
		MMF-Net                 & 2×R50          & —      & 45.5    & — \\
		RedNet                  & 2×R50          & —      & 47.8    & 22.1 \\
		RDFNet~\cite{park2017rdfnet}                  & 2×R50          & 47.7*   & —       & 7.2 \\
		ACNet~\cite{hu2019acnet}                   & 3×R50          & 48.3   & 48.1    & 16.5 \\
		SA-Gate~\cite{chen2020bi}                 & 2×R50          & 50.4   & 49.4*   & 11.9 \\
		\hline
		SGNet~\cite{chen2021spatial}                   & R101           & 49.0   & 47.1    & — $\nabla$ \\
		Idempotent              & 2×R101         & 49.9   & 47.6    & — $\nabla$ \\
		2.5D Conv               & R101           & 48.5   & 48.2    & — $\nabla$ \\
		\hline
		MMF-Net                 & 2×R152         & 44.8   & 47.0    & — $\nabla$ \\
		RDFNet                  & 2×R152         & 50.1*  & 47.7*   & 5.8 \\
		\hline
		ESANet-R18              & 2×R18          & 47.3   & 46.2    & 34.7 \\
		ESANet-R18-Nbt1D        & 2×R18 Nbt1D    & 48.2   & 46.9    & 36.3 \\
		ESANet-R34              & 2×R34          & 48.8   & 47.1    & 27.5 \\
		\textbf{ESANet-R34-Nbt1D} & \textbf{2×R34 Nbt1D} & \textbf{50.3} & \textbf{48.2} & \textbf{29.7} \\
		ESANet-R50              & 2×R50          & 50.5   & 48.3    & 22.6 \\
		\hline
		ESANet (pretrain)       & 2×R34 Nbt1D    & 51.6   & 48.0    & 29.7 \\
		\hline
	\end{tabular}
	
	\caption{Quantitative comparison of ESANet with state-of-the-art RGB-D semantic segmentation methods on NYUv2 and SUNRGB-D benchmarks, adopted from Table~I of ESANet~\cite{seichter2021efficient}. Results demonstrate ESANet's superior trade-off between segmentation accuracy and inference speed on NVIDIA Jetson AGX Xavier. Legend: R: ResNet backbone, *: additional test-time augmentation (not timed), N/A: no implementation available, $\dagger$: includes operations unsupported by TensorRT, and $\nabla$: slower expected due to complex backbone. This supports ESANet's suitability for real-time (online) semantic segmentation in resource-constrained robotic systems.}
	\label{tab:segmentation_summary}
\end{table}

\subsection{Our Contribution in Context}
Our review of the literature reveals a clear and critical gap. While classical planners are rigorous but semantically blind, modern learning-based planners are intelligent but computationally expensive and cannot guarantee adherence to explicit, user-defined rules. This leaves a need for a hybrid framework that is simultaneously \textbf{(1) cost-effective}, replacing LiDARs with low-cost cameras; \textbf{(2) computationally efficient}, running in real-time on embedded hardware; \textbf{(3) adaptable}, operating in unknown environments using a dynamic occupancy grid; and \textbf{(4) semantically aware}, able to interpret and strictly enforce user-defined, non-geometric constraints.

Our work is the first to address this four-fold challenge. We propose a framework that tightly integrates the efficient ESANet semantic segmentation model with a dynamic, grid-based online A* path planner, delivering a practical and robust solution that combines the best of both classical and modern paradigms.

\begin{figure}[t!]
	\centering
	\includegraphics[width=1.0\columnwidth, trim={5.5cm 4.5cm 5.0cm 5.0cm}, clip]{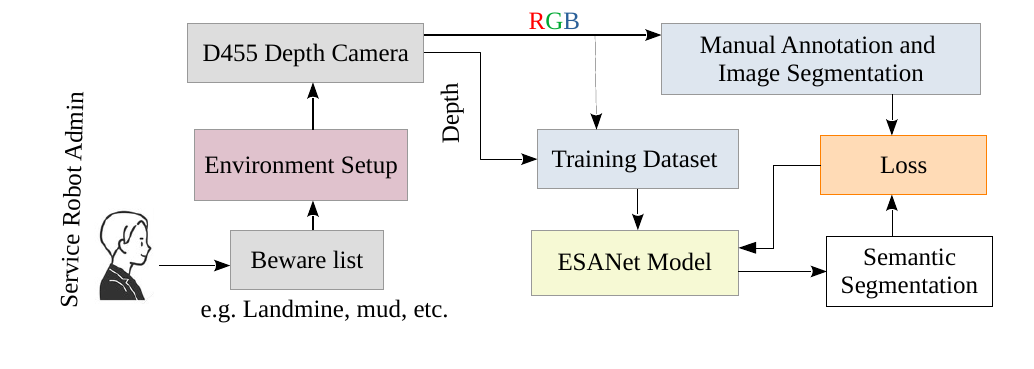}
	\caption{Perception Module Training Pipeline: The process begins with the administrator defining a workspace-specific "Beware list" and setting up the environment. The D455 camera captures RGB and Depth data. RGB images are manually annotated based on the "Beware list" to create a labeled training dataset. This dataset (RGB images, Depth maps, and Semantic Labels) is used to train the ESANet model via supervised learning, optimizing a loss function to produce an accurate semantic segmentation model tailored for the target application.}
	\label{fig:kgraph}
\end{figure}

\section{METHODOLOGY}
The workspace is defined as $P \subset \mathbb{R}^3$, which is unknown or partially known to the wheeled robot during navigation. The Depth Camera mounted on the robot captures a region $Q \subset P$, which lies within its field of view (FoV). The valid data from the camera sensor is represented as $B \subset Q$, constrained by the sensor's depth-sensing limits. The robot, navigating in the x-y plane, classifies an object as an obstacle if its height exceeds 10 cm. The set of obstacles, denoted as $O \subset B$, consists of all objects in B whose height exceeds this threshold. The set $K = B - O$ represents the visual information which may be relevant for safe path planning. The set $K$ (RGB-D data) is first processed to identify the relevant data (e.g. robot safety critical, an environment and application specific) $D \subset K$ (segmented output) based on a beware list (user-centric) within the specific workspace. The data $D$ is converted into the 3D pointcloud using depth frame and then adds with filtered pointcloud obtained from data $O$. These summed pointcloud is then converted into occupancy grid. Then, the A* path finding algorithm is applied to generate a collision-free path. This entire process from capturing depth and RGB data to path planning is repeated for each new camera frame until the robot reaches its target.

\begin{algorithm}[!t]
	\caption{\textbf{Vision-Aided Online Map Planning and Path Generation}}
	\label{alg:vision_online_planning}
	\begin{algorithmic}[1]
		\Require
		Incoming point cloud $P_t=\{x,y,z\}$, odometry $O_t=(x_r,y_r,\theta_r)$, goal $G(x_g,y_g)$, grid resolution $r_g$, 
		grid size $(W,H)$, vehicle width $w_v$, safety margin $s_m$
		\Ensure
		Collision-free path $\Pi=\{p_1,p_2,\dots,p_n\}$ and navigable waypoints $W_p=\{w_1,w_2,\dots,w_m\}$
		
		\State Initialize occupancy grid $M(W,H)\gets0$
		\For{each incoming point cloud frame $P_t$}
		\State Project $(x,y)$ to map coordinates: 
		$(u,v)=\lfloor(x,y)/r_g\rfloor+(W,H)/2$
		\State Mark occupied cells: $M(v,u)\gets255$
		\EndFor
		\State Continuously update odometry $O_t=(x_r,y_r,\theta_r)$ and wait for goal $G(x_g,y_g)$
		
		\Statex
		\State \textbf{Map Preprocessing:}
		\State Copy map $M'\gets M$
		\State Compute inflation radius $r_i=\left\lceil\frac{(w_v/2+s_m)}{r_g}\right\rceil$
		\State Dilate $M'$ using circular kernel of radius $r_i$
		
		\Statex
		\State \textbf{Path Planning (A$^*$\,Algorithm):}
		\State Convert start and goal into pixel coordinates:
		$S=\text{world\_to\_pixel}(x_r,y_r)$, 
		$G=\text{world\_to\_pixel}(x_g,y_g)$
		\State Initialize open set $O\gets\{S\}$, cost map $g(S)\gets0$, parent map $C\gets\emptyset$
		\While{$O\neq\emptyset$}
		\State Expand node $n$ with minimum $f(n)=g(n)+h(n,G)$
		\For{each 8-connected neighbor $n'$ of $n$}
		\If{$n'$ free in $M'$ and not in $C$}
		\State Update $g(n')$, add to $O$, set $C(n')\gets n$
		\EndIf
		\EndFor
		\If{$n==G$}
		\State \textbf{break}
		\EndIf
		\EndWhile
		\State Reconstruct path $\Pi$ by backtracking via $C$
		
		\Statex
		\State \textbf{Path Refinement:}
		\For{each consecutive triplet $(p_{i-1},p_i,p_{i+1})$ in $\Pi$}
		\State Compute turning angle 
		$\theta=\cos^{-1}\!\!\left(\dfrac{v_1\cdot v_2}{\|v_1\|\|v_2\|}\right)$
		\If{$\theta<\theta_{th}$}
		\State Keep $p_i$ as a waypoint
		\EndIf
		\EndFor
		\State Convert all selected points to world coordinates
		\State Publish $\Pi$ and $W_p$ for navigation
	\end{algorithmic}
\end{algorithm}

\subsection{Integrated Real-Time Perception and Mapping}
Our robust, multi-layered architecture for autonomous mobile robot navigation processes online RGB-D frames to safely and efficiently guide service robots to their goal in unknown or partially known, dense, and cluttered environments, as conceptually illustrated by Fig.~\ref{fig:rednet}. The system begins with real-time semantic perception: ESANet performs binary semantic segmentation to identify user-centric and critical environment constraints. This semantically tagged data is instantly converted to a 3D point cloud, merged with raw depth data pre-filtered to eliminate irrelevant points (ground plane and high-altitude features), and then continuously integrated into the explored Global 3D Map. This map is projected onto the 2D navigation surface to form a dynamic Occupancy Grid (initialized as unknown at t=0). Following each update, the A* pathplanning algorithm rapidly computes the shortest, globally collision-free path, which is then decomposed into local waypoints based on path curvature. These waypoints guide the Local Path Planner, built on motion primitives, which performs the critical high-frequency tasks of trajectory generation, real-time obstacle avoidance, dynamic environment handling, and global plan following. The entire process, enhanced by a modified terrain analysis method~\cite{cao2022autonomous}, culminates in the generation of the necessary linear and angular velocities required for the service robot's safe and efficient traversal.\\

\subsection{3D Point Generation from Segmented Image}
To obtain a 3D point cloud of the segmented foreground (Beware list represented by the set $c$) region, we process a depth frame $D \in \mathbb{R}^{H \times W}$ and binary segmented image $S \in \mathbb{N}^{H \times W}$ where $S(i,j) \in \{0, 1\}$. First, we define a binary mask:

\begin{equation}
M_c(i,j) = 
\begin{cases} 
1 & \text{if } S(i,j) = c \\
0 & \text{otherwise}
\end{cases}
\label{eq:mask}
\end{equation}

The foreground depth map is then obtained through element-wise multiplication, $D_c(i,j) = D(i,j) \cdot M_c(i,j)$. Using camera intrinsic parameters given focal lengths $(f_x, f_y)$ and optical center $(c_x, c_y)$, the 3D coordinate of each active pixel $(i,j)$ (where $M_c(i,j)=1$) is

\begin{equation}
\begin{aligned}
Z_c &= D_c(i,j) \\
X_c &= \frac{(j - c_x)Z_c}{f_x} \\
Y_c &= \frac{(i - c_y)Z_c}{f_y}
\end{aligned}
\label{eq:3d_projection}
\end{equation}

This projection can alternatively be expressed in matrix form using the inverse intrinsic matrix $K^{-1}$:
\begin{equation}
\begin{bmatrix}
X_c \\ Y_c \\ Z_c
\end{bmatrix}
= D_c(i,j) \cdot K^{-1}
\begin{bmatrix}
j \\ i \\ 1
\end{bmatrix}
\label{eq:matrix_projection}
\end{equation}

The collection of all $\mathbf{P}_{3D}(i,j) = [X_c, Y_c, Z_c]^\top$ forms the 3D point cloud of the segmented foreground in camera coordinates. These points are immediately added to the navigation map so that both the global path planner and the local trajectory planner treat them as objects to be respected and avoided.
  
\subsection{Online Exploration and Global Mapping}
The global occupancy grid is constructed by continuously integrating point clouds generated from real-time RGB-D data. Each incoming RGB-D frame (color and depth) is processed to extract 3D points, which are then transformed into the global coordinate frame. To minimize computational overhead and reduce drift caused by sensor noise (from the Intel RealSense D455, wheel encoders) and robot skidding, our system dynamically updates the global map by replacing outdated point clouds with newly captured data.
The proposed framework processes each RGB-D frame along with the corresponding robot pose to generate a combined point cloud associated with a unique pose ID. Initially, the global map is empty; the first point cloud is added and stored with its pose ID. For each subsequent update, the system checks whether a point cloud with the same pose ID already exists in the global map. If it does, the old data is replaced with the new point cloud; otherwise, the new data is appended. This selective replacement strategy ensures efficient memory usage, reduces redundancy, and maintains an accurate and up-to-date global representation—making it highly suitable for real-time service robot navigation.

\begin{figure*}[t!]
	\centering
	\includegraphics[width=2.0\columnwidth, trim={3.0cm 10.0cm 3.0cm 2.0cm}]{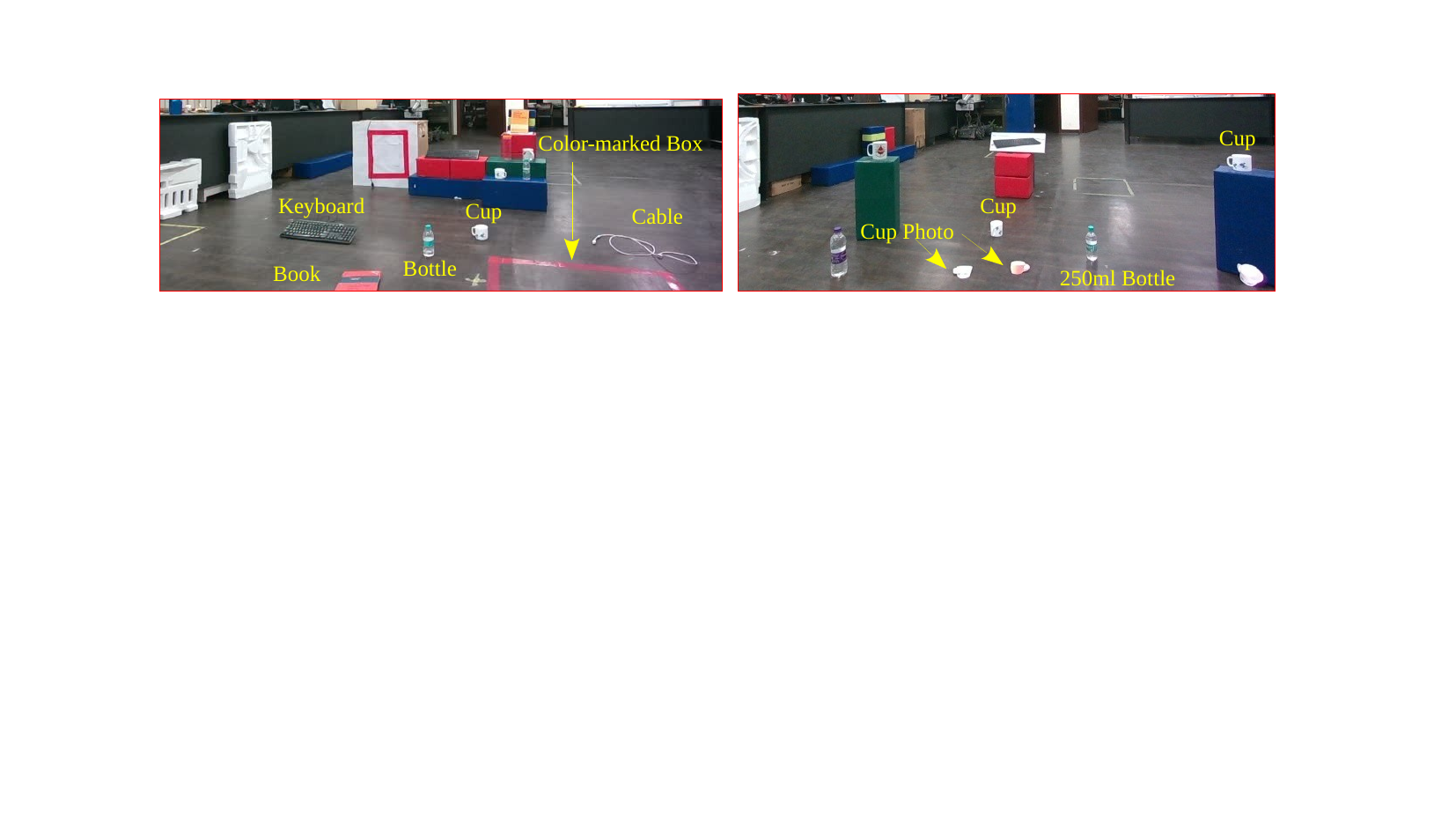}
	\caption{Navigation workspace populated with user-defined items and color-marked zones placed on the ground surface. Left and right frames illustrate the scene setup, highlighting areas that the robot must detect and avoid during path planning.}
	\label{fig:scene}
\end{figure*}
\subsection{Dynamic Path Planning}
The robot's mission is to find the shortest, collision-free path to its destination while strictly avoiding all obstacles, color-marked prohibited zones, and specific objects defined in a beware list. As the robot moves and explores the environment online, it constantly gathers new information, requiring the path to be dynamically updated.

To handle this challenge, we select $A^*$ as our foundational planning algorithm. While incremental algorithms like $D^*$~\cite{stentz1994optimal} and $D^*$ Lite are popular for efficiently replanning by reusing previous path segments, they treat the start point as fixed, leading to unnecessary computation on already traversed path segments and complicating real-time tracking of the robot's dynamic position. Instead, our method opts for full $A^*$ replanning: whenever new environmental information is detected, the algorithm is executed from the robot's current position to the fixed destination, treating all unexplored areas as free space. This simple, effective approach provides a clean, optimal path without the overhead of maintaining and pruning covered sections.

We exclude other common approaches for the following reasons: Learning-based planners are unsuitable for unknown environments, as their performance drops significantly outside their trained areas. Random sampling-based planners (like RRT and PRM~\cite{kavraki2002probabilistic} and their variants~\cite{rajasa2024smart}~\cite{novosad2023ctopprm}) are excellent for free space but often suffer from long planning times and high computational demands in unknown territories due to redundant sampling and the cost of updating the search graph as the environment is revealed. Our chosen $A^*$ replanning strategy provides the best balance of computational simplicity, guaranteed optimality, and fast responsiveness for effective online navigation.\\

\begin{algorithm}
	\caption{Training and Inference with ESANet for Binary Semantic Segmentation and 3D Point Cloud Fusion}
	\begin{algorithmic}[1]
		
		\State \textbf{Input:} Dataset $\mathcal{D} = \{(RGB_i, Depth_i, GT_i)\}$, D455 camera parameters
		\State \textbf{Output:} Composite 3D point cloud with filtered foreground objects
		
		\Procedure{TrainESANet}{$\mathcal{D}$}
		\State Initialize ESANet model with ResNet18 backbone for binary segmentation
		\State Define loss function $\mathcal{L}$ and optimizer
		\For{epoch in training epochs}
		\For{batch $(RGB_b, Depth_b, GT_b)$ in $\mathcal{D}$}
		\State $Pred_b \gets$ ESANet($RGB_b$, $Depth_b$)
		\State $loss \gets \mathcal{L}(Pred_b, GT_b)$
		\State Backpropagate $loss$ and update model params
		\EndFor
		\State Validate model performance
		\EndFor
		\EndProcedure
		
		\Procedure{SegAndPC}{$RGB_{frame}, Depth_{frame}$}
		\State $SegMask \gets$ ESANet($RGB_{frame}, Depth_{frame}$)
		\State $FGPixels \gets \{p | SegMask(p) = 1\}$
		\State $FGPoints \gets \emptyset$
		\For{each pixel $p$ in $FGPixels$}
		\State $d \gets Depth_{frame}(p)$
		\State $pt_{3D} \gets$ PixelToWorld($p$, $d$, camera params)
		\State $FGPoints \gets FGPoints \cup \{pt_{3D}\}$
		\EndFor
		\State $FullPC \gets$ ConvertDepthToPointCloud($Depth_{frame}$)
		\State $CompositePC \gets FullPC \cup FGPoints$
		\State \Return $CompositePC$
		\EndProcedure
		
	\end{algorithmic}
\end{algorithm}
\subsection{Local and Global Path Planners} The dynamic path-finding architecture is specifically designed to safely and efficiently navigate unknown and partially known environments toward a fixed destination. This is achieved by structuring the system around two distinct, yet highly coordinated, components: the Global Planner and the Local Planner. The Global Planner operates on a dynamically updated global map, which integrates explored areas with unexplored regions (treated as free space). Utilizing the $A^*$ algorithm, it computes the high-level, shortest collision-free path and identifies crucial intermediate waypoints that guide the robot's next local goal. The Local Planner, in turn, uses the waypoint and terrain data (within the waypoint's radius) to rapidly select a detailed, real-time feasible trajectory from a predefined set of options. It utilizes a motion-primitive-based approach (such as the method detailed in Falco~\cite{zhang2020falco}) to rapidly evaluate and select safe, kinodynamically feasible trajectories that respect the robot’s dynamics and immediately adapt to local environmental constraints. Ultimately, the Local Planner provides the necessary linear and angular velocities to the Husky A200 mobile robot platform (a differential drive robot), allowing the Motion Control Unit (MCU) to control the speed of the left and right wheel pairs, thereby ensuring effective and safe movement on the ground.

\section{EXPERIMENTS}
Our proposed method was evaluated in both simulation and real-world environments to validate its
effectiveness in navigating complex scenarios incorporating both geometric and semantic constraints.
Simulation experiments were conducted on a system powered by an Intel(R) Core(TM) i7-4770 CPU
running at 3.40GHz, with 8 cores and 2 threads per core, utilizing Gazebo, a high-fidelity 3D robotics
simulator, for virtual testing. Real-world tests were performed on embedded hardware, detailed below.

\begin{figure*}[t!]
	\centering
	\includegraphics[width=1.9\columnwidth, trim={3.0cm 2.0cm 3.0cm 2.0cm}]{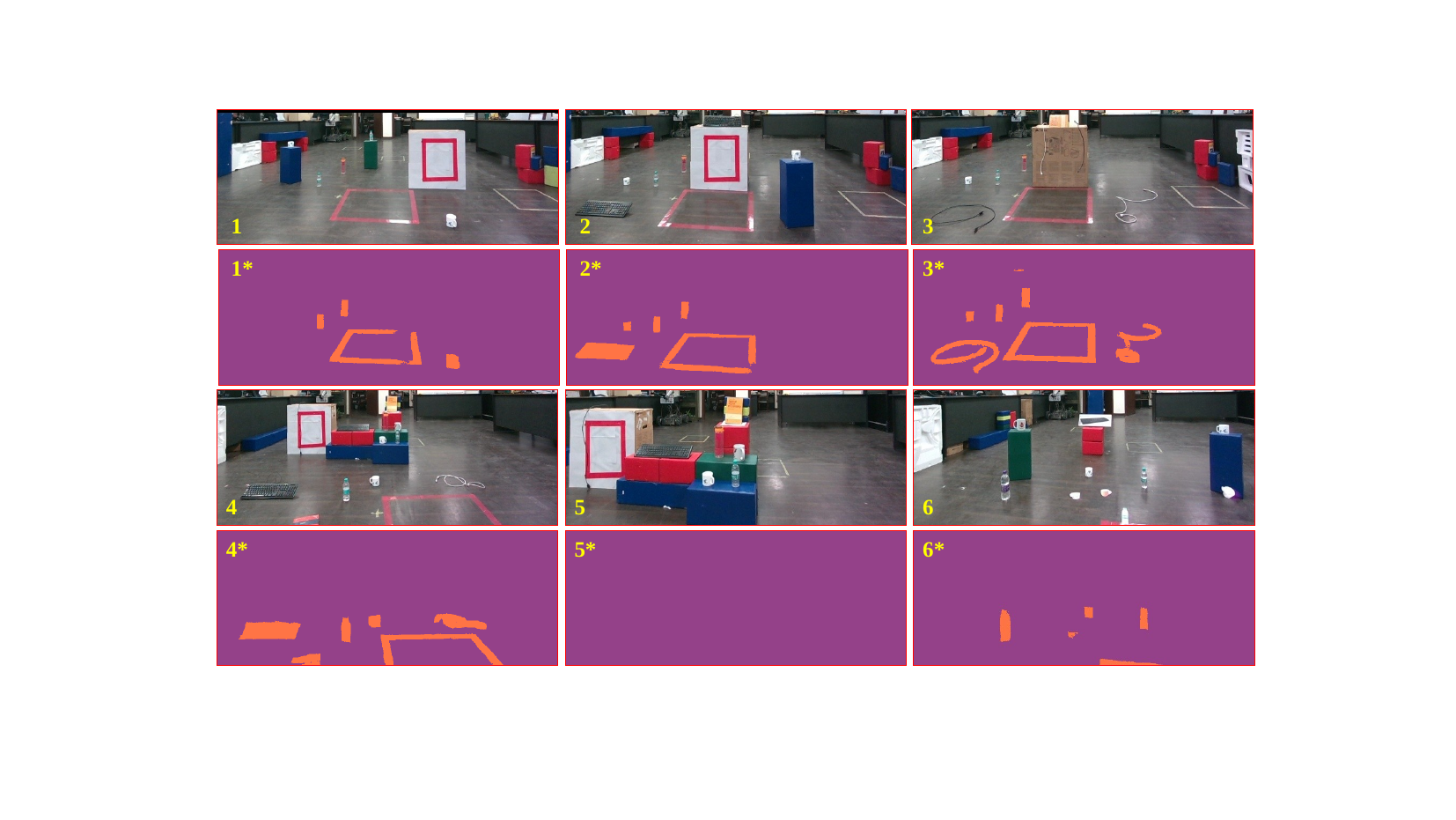}
	\caption{RGB-D Semantic Segmentation Results. Frames 1 to 6 display a workstation navigation scene containing both considerable and nonconsiderable objects. Frames $1^*$ to $6^*$ show the corresponding semantic segmentation outputs generated by our trained ESANet model. These masks highlight the identified "Beware list" items (e.g., objects to avoid, designated zones), demonstrating the model's ability to accurately perceive and classify task-relevant semantic information from raw RGB input.}
	\label{fig:results}
\end{figure*}
\subsection{Simulation Results}
We designed two primary simulation scenarios within Gazebo to rigorously test the framework’s capa-
bilities across different environmental complexities and constraint types, focusing first on the perception
module and then on integrated navigation.
Perception Performance: The foundation of our approach lies in accurate semantic perception.
Fig.~\ref{fig:results} showcases representative outputs of the trained ESANet model operating on RGB-D data
from a typical indoor scene. Frames 1-6 display the input RGB images containing various objects and
environmental features, including items designated as critical (e.g., objects on the floor, red-marked
zones) and non-critical clutter. The corresponding frames 1*-6* show the binary segmentation masks
generated by ESANet. These masks accurately isolate pixels belonging to objects or regions defined in
the ``Beware list,'' demonstrating the perception module’s capability to extract the specific task-relevant
semantic information required by the planner from the raw visual input, effectively filtering out irrelevant
elements.
\begin{figure*}[t!]
	\centering
	\begin{subfigure}{0.67\columnwidth}
		\centering
		\includegraphics[trim={7cm 1.5cm 6.0cm 0.0cm}, clip, width=\linewidth]{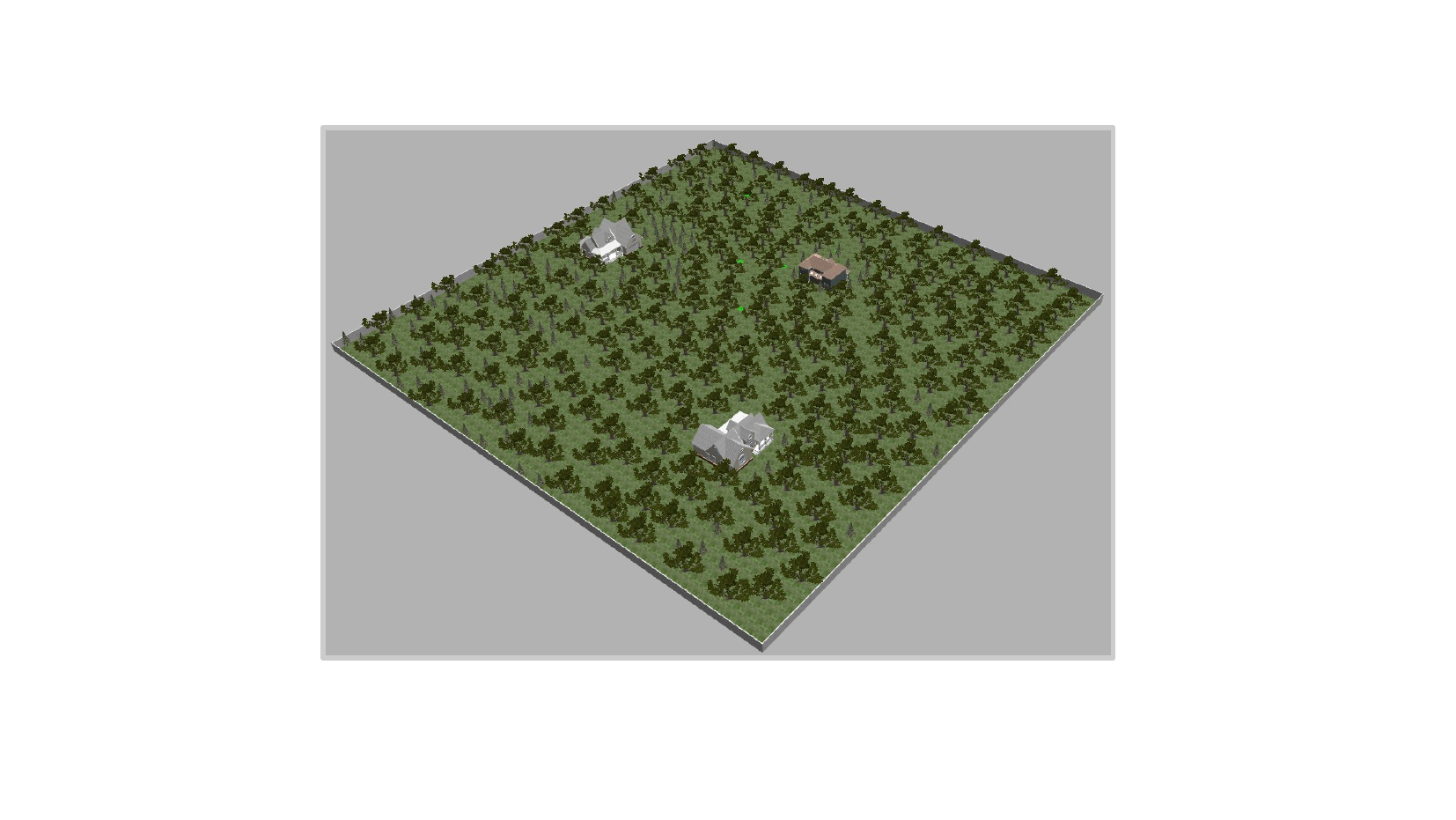}
	\end{subfigure}
	\hfill
	\begin{subfigure}{0.75\columnwidth}
		\centering
		\includegraphics[trim={6.0cm 0.0cm 5.8cm 0.0cm}, clip, width=\linewidth]{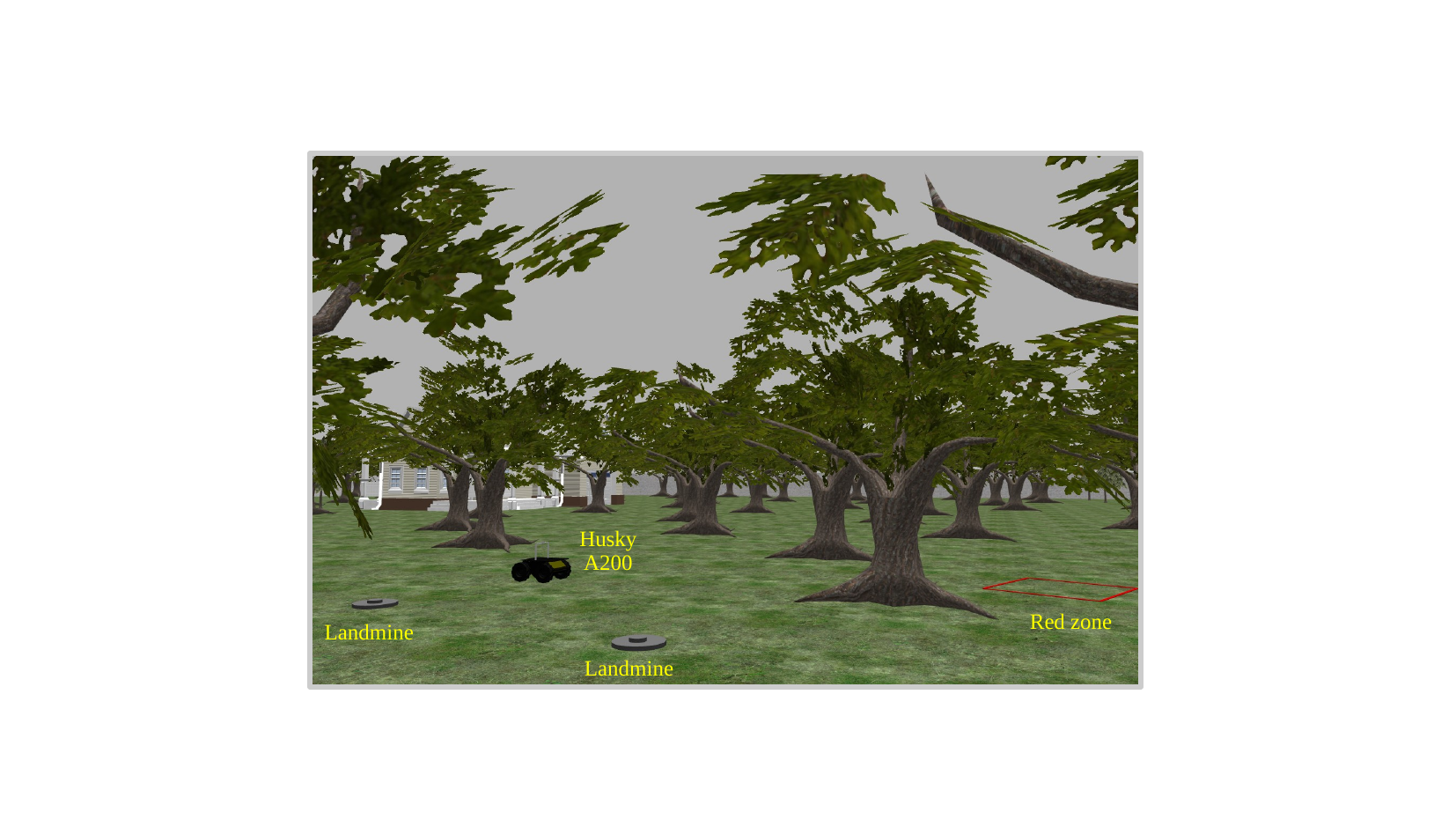}
	\end{subfigure}
	\hfill
	\begin{subfigure}{0.58\columnwidth}
		\centering
		\includegraphics[trim={9cm 3.5cm 8.0cm 3.0cm}, clip, width=\linewidth]{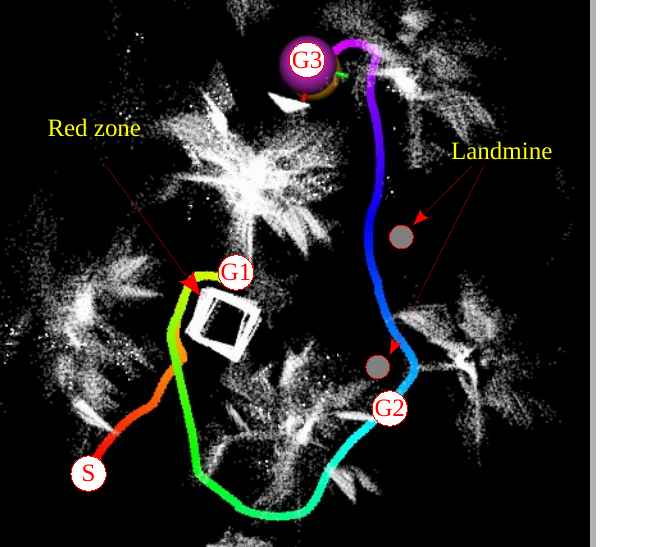}
	\end{subfigure}
	    \\[4pt]  
	\hspace{0.5cm}(a)\hspace{6.8cm}(b)\hspace{5.8cm}(c)
	\caption{Simulation in a Forest Environment with Semantic Hazards. (a) Top-down view of the Gazebo forest environment (150m x 150m), featuring scattered buildings, numerous trees, and designated hazardous zones (landmines and a red zone). (b) Depiction of the Husky A200 robot within the forest environment, showing the detected ``Landmine'' objects and a ``Red Zone'' in the foreground, as perceived by the onboard RGB-D camera. This perspective highlights the challenge of navigating amidst dense vegetation while accurately identifying and avoiding critical semantic obstacles. (c) Overhead visualization of the robot's exploration and navigation trajectory (multi-color line) from Start (S) through intermediate goals (G1, G2) to the final destination (G3). The white points represent the mapped environment. The path successfully navigates around both the geometrically complex trees and the semantically defined ``Landmine'' and ``Red Zone'' hazards, demonstrating the framework's capability to integrate diverse constraint types.}
	\label{fig6}
\end{figure*}
\begin{figure}[!t]
	\centering
	\includegraphics[width=1.0\columnwidth, trim={3.0cm 0.5cm 3.0cm 1.2cm}]{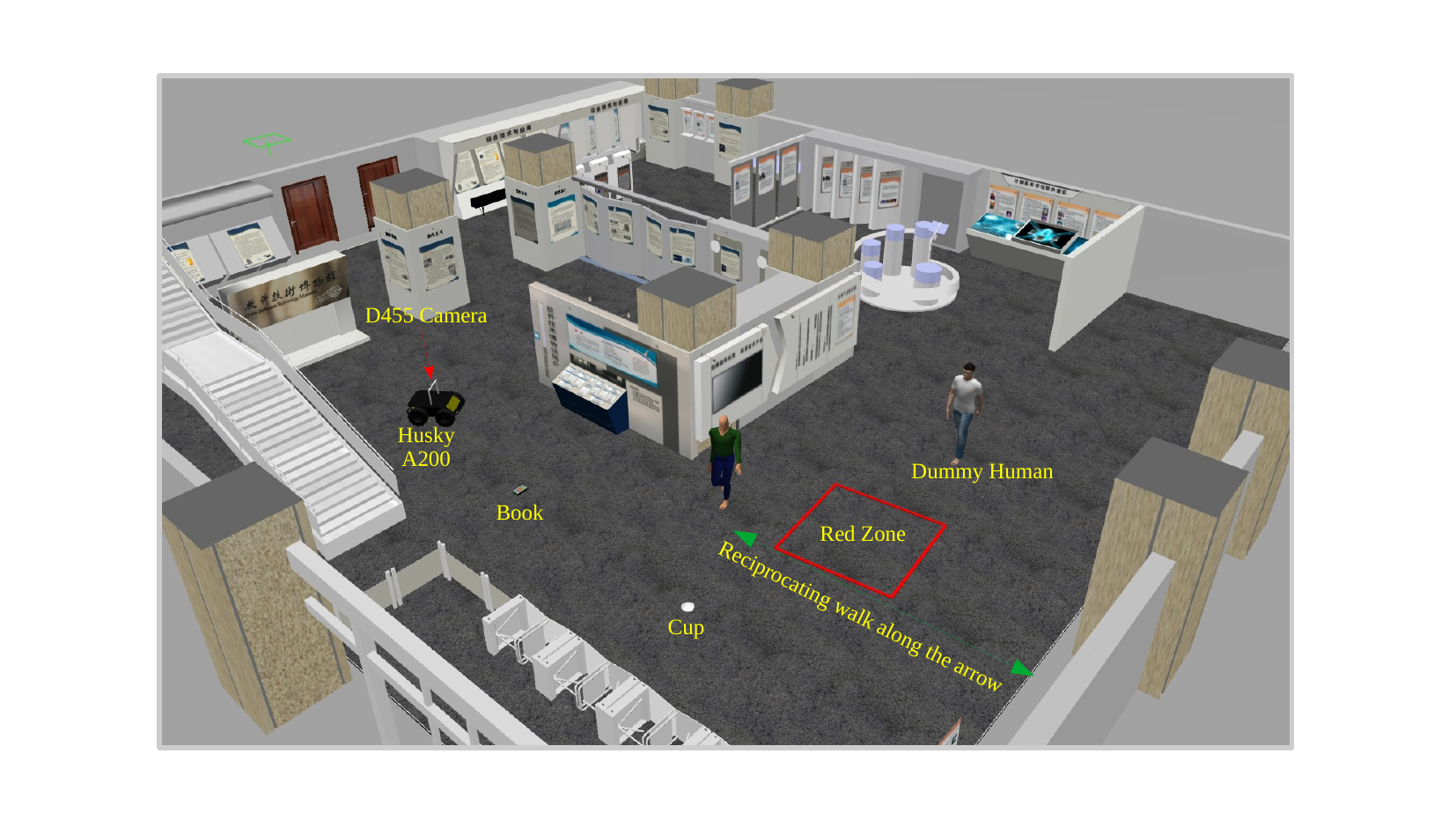}
	\caption{Gazebo Simulation Environment Setup. The scene includes a Husky A200 robot equipped with a D455 depth camera, static obstacles, user-defined semantic obstacles (a book and a cup), a non-geometric constraint (red zone), a dummy human  and a dynamic agent (moving along the indicated path).}
	\label{simSetup}
\end{figure}
\begin{figure}[!t]
	\centering
	\includegraphics[width=1.0\columnwidth,trim={8.0cm 2.0cm 8.0cm 0.8cm}]{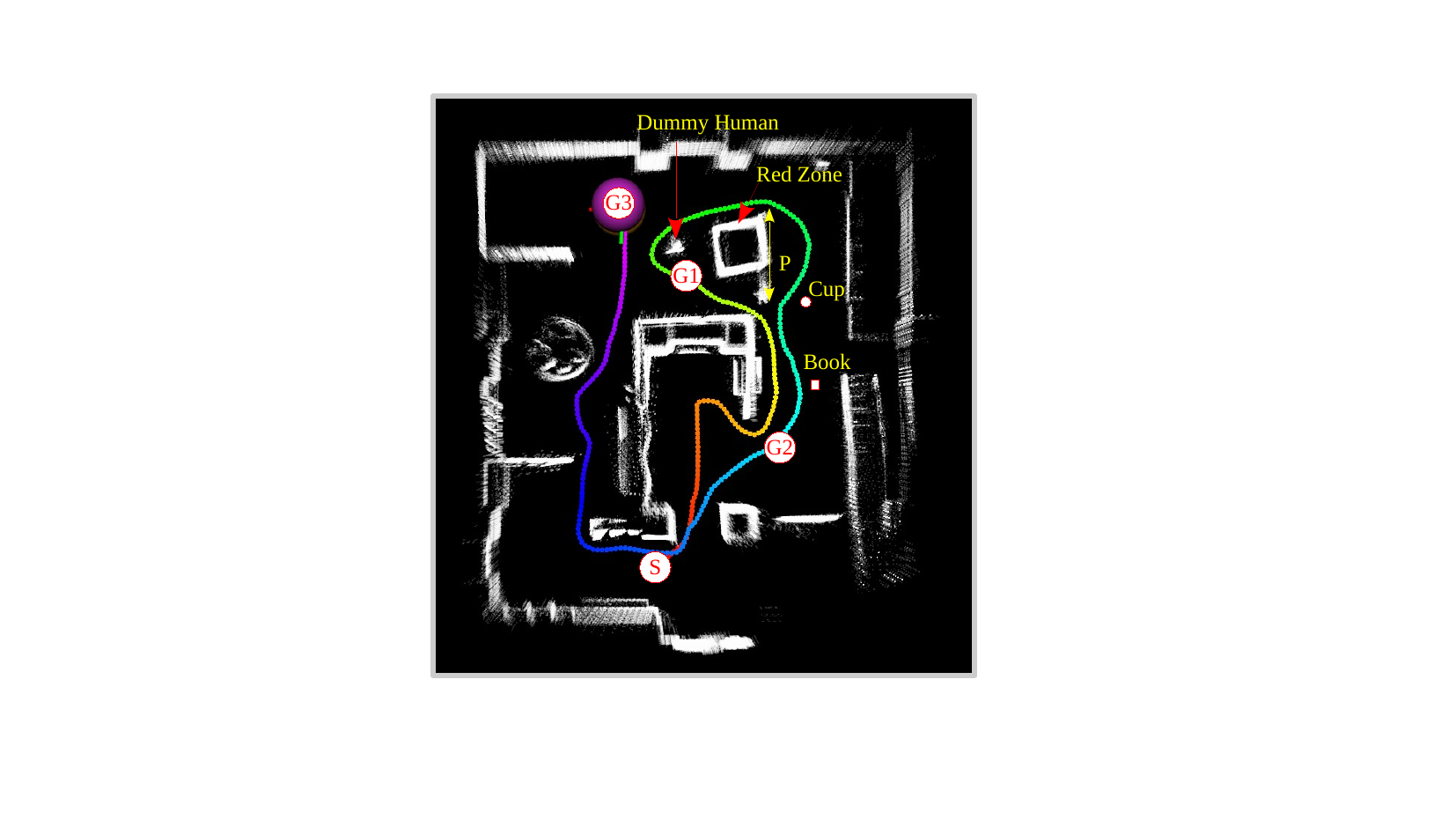}
	\caption{Top-down View of a Multi-Goal Simulation Trajectory. This map shows the environment (represented by the point cloud) explored by the robot and the complete path executed (multi-color line). The robot navigated sequentially from the Start (S) to Goal 1 (G1), then to Goal 2 (G2), and finally to Goal 3 (G3). The path demonstrates successful navigation around the dynamic human agent (path indicated by the yellow arrow), the static semantic obstacles (Cup, Book), and the non-geometric constraint (Red Zone) while achieving the sequence of specified goals.}
	\label{topView}
\end{figure}

\begin{figure}[!t]
	\centering
	\includegraphics[width=1.0\columnwidth, trim={8.0cm 4.0cm 8.0cm 4.0cm}, clip]{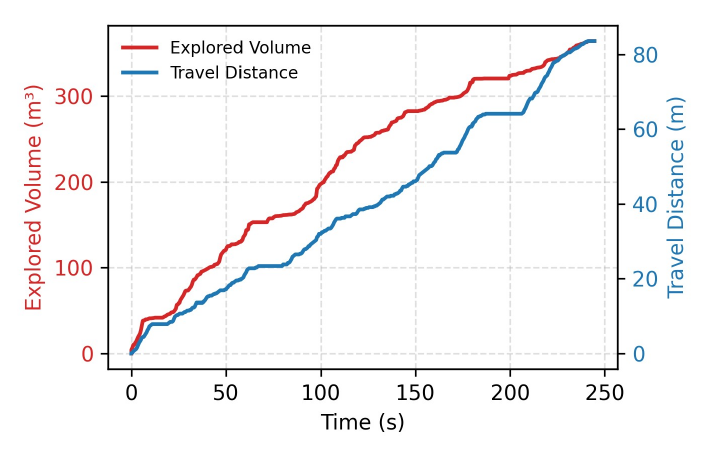}
	\caption{Exploration and navigation performance in the dense simulated environment shown in~\ref{simSetup}. The results highlight the capability of the proposed algorithm to navigate efficiently while avoiding geometric and non-geometric obstacles.}
	\label{simgraph}
\end{figure}
Forest Scenario: To evaluate robustness in geometrically complex environments with severe seman-
tic hazards, we utilized a simulated forest setting (150m × 150m), depicted in Fig.~\ref{fig6}. This scenario
featured dense trees and scattered buildings (geometric obstacles), along with hazardous semantic items
(``Landmine'') placed on the ground and a designated ``Red Zone'' (Fig.~\ref{fig6}a). Fig.~\ref{fig6}b provides the
robot’s first-person perspective within this challenging environment, highlighting the visual complexity
and the difficulty of reliably detecting semantic ground hazards amidst natural clutter. Fig.~\ref{fig6}c shows
the resulting top-down map (white points indicate explored obstacles/terrain) and the successfully executed trajectory (multi-color line) from Start (S) to Goal (G3) via intermediate waypoints (G1, G2).
The path demonstrates the system’s capability to navigate the intricate arrangement of trees while cor-
rectly identifying and circumventing both the discrete ``Landmine'' objects and the extended ``Red Zone,''
validating its effectiveness in integrating diverse constraint types within complex geometries.
Indoor Scenario: As illustrated in Fig.~\ref{simSetup}, we also constructed a realistic indoor environment
(approximately 100m × 100m) populated with typical furniture and structures acting as static geometric
obstacles. Crucially, the scene included user-defined semantic elements: physical objects designated
as critical via the ``Beware list'' (a book and a cup placed on the floor), a non-geometric constraint
represented by a red square zone marked on the floor, a stationary dummy human, and a dynamic human
agent programmed to walk repeatedly along a path intersecting the robot’s potential routes (indicated
by the green arrow). A Husky A200 robot model equipped with a simulated D455 depth camera served
as the navigation agent. This setup allowed us to evaluate the system’s integrated ability to handle
static and dynamic obstacles, geometric barriers, and both object-based and area-based semantic rules
simultaneously.
Fig.~\ref{topView} presents a top-down visualization of the map generated and the path executed during a
successful multi-goal navigation task within this indoor environment. The robot was tasked to navigate
sequentially from a starting point (S) to three distinct goal locations (G1, G2, G3). The generated
trajectory (multi-color line) demonstrates the successful avoidance of all constraint types: the static
geometric obstacles (walls, furniture represented in the point cloud map), the dynamic human agent
(whose path is indicated by the yellow arrow, P), the static semantic obstacles (``Cup'', ``Book''), and
the non-geometric constraint (``Red Zone''). The robot efficiently reached all specified goals in the
correct sequence, validating the core functionality of the hybrid planner. Further illustrating the system’s
performance, Fig.~\ref{simgraph} plots the explored volume and total travel distance against time for a typical run
in this environment, showing efficient exploration and movement while maintaining safety.
\begin{figure}[!t]
	\centering
	\includegraphics[width=1.0\columnwidth, trim={3.0cm 3.0cm 1.0cm 2.0cm}, clip]{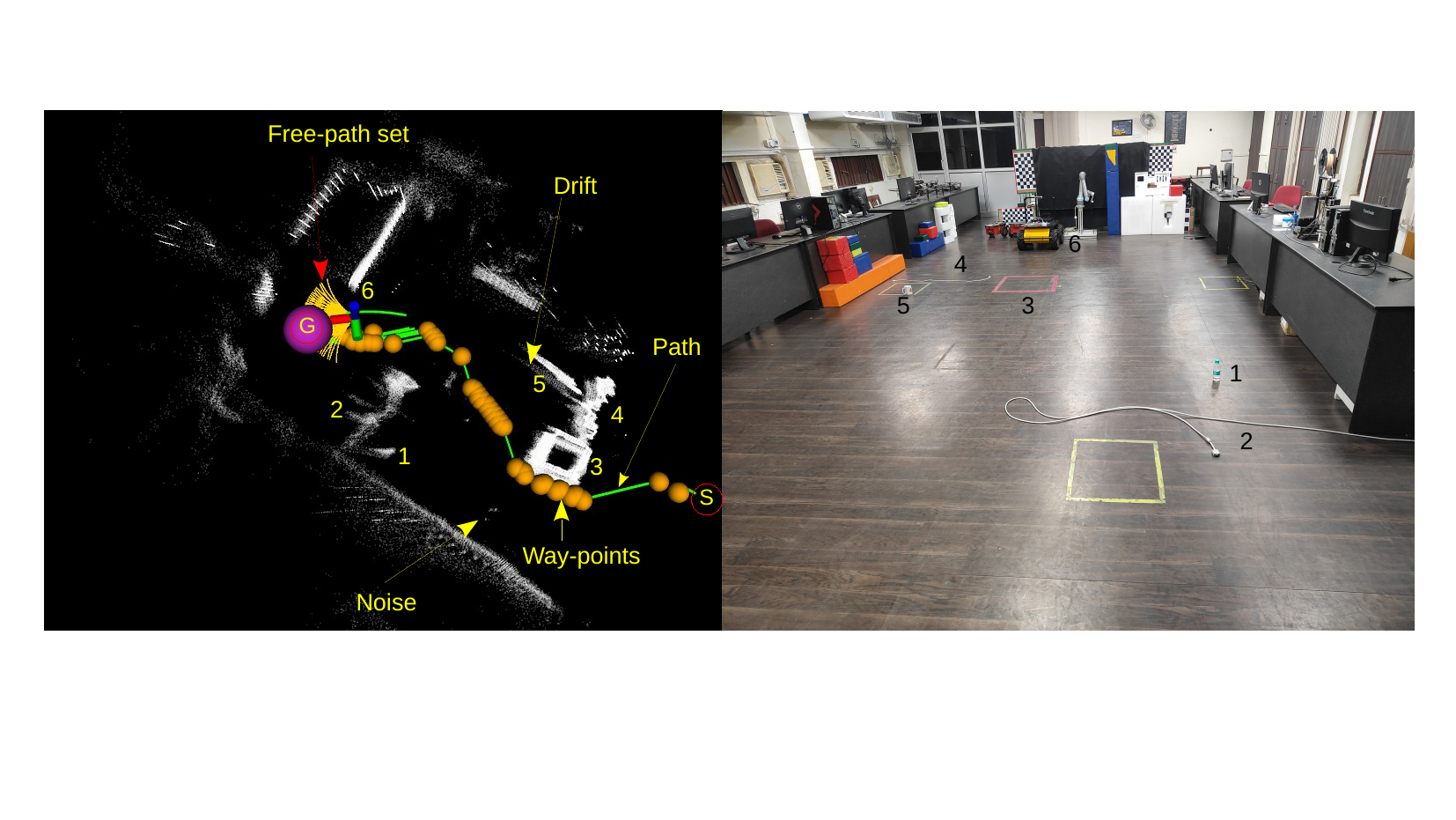}
	\caption{Collision-Free Traversal and Global Mapping Performance: This figure displays the Husky A200 robot's navigation task, mapping the real-world environment (Right, RGB image) to its localization data (Left, fused path and point cloud). The setup includes geometric obstacles and five user-defined hazards (1. bottle, 2. cable, 3. box, 4. cable, 5. cup). The left side demonstrates the algorithm's performance, showing the initial planned path, calculated way-points, and the final collision-free traversed path achieved through exploration. The visualization also clearly illustrates localization components such as minimal noise and limited drift from the start/goal points. }
	\label{rw1}
\end{figure}
\begin{figure*}[!t]
	\centering
	\includegraphics[width=2.0\columnwidth, trim={1.5cm 6.5cm 1.5cm 4.5cm}, clip]{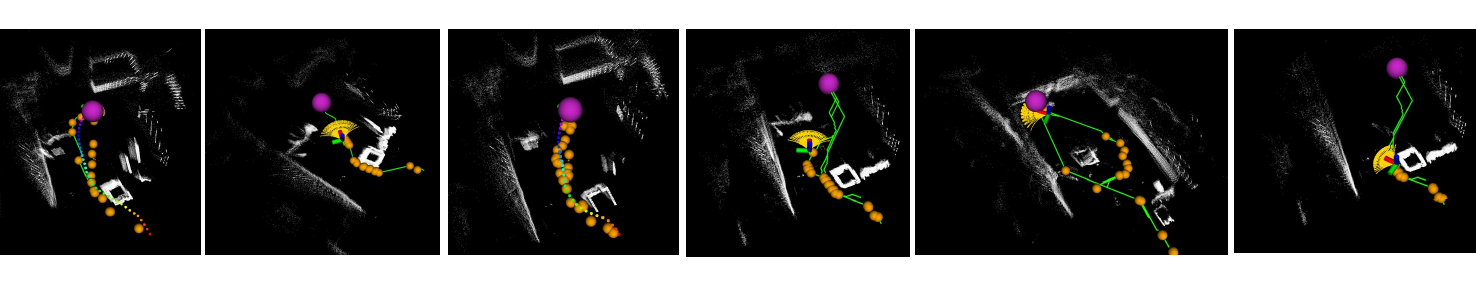}
	\caption{Point cloud views of the real-world environment illustrating successful path planning and traversal across multiple goal points, with our algorithm avoiding both obstacles and user-specified restricted areas.}
	\label{rw2}
\end{figure*}
\begin{figure}[!t]
	\centering
	\includegraphics[width=1.0\columnwidth, trim={3.0cm 3.0cm 1.0cm 2.0cm}, clip]{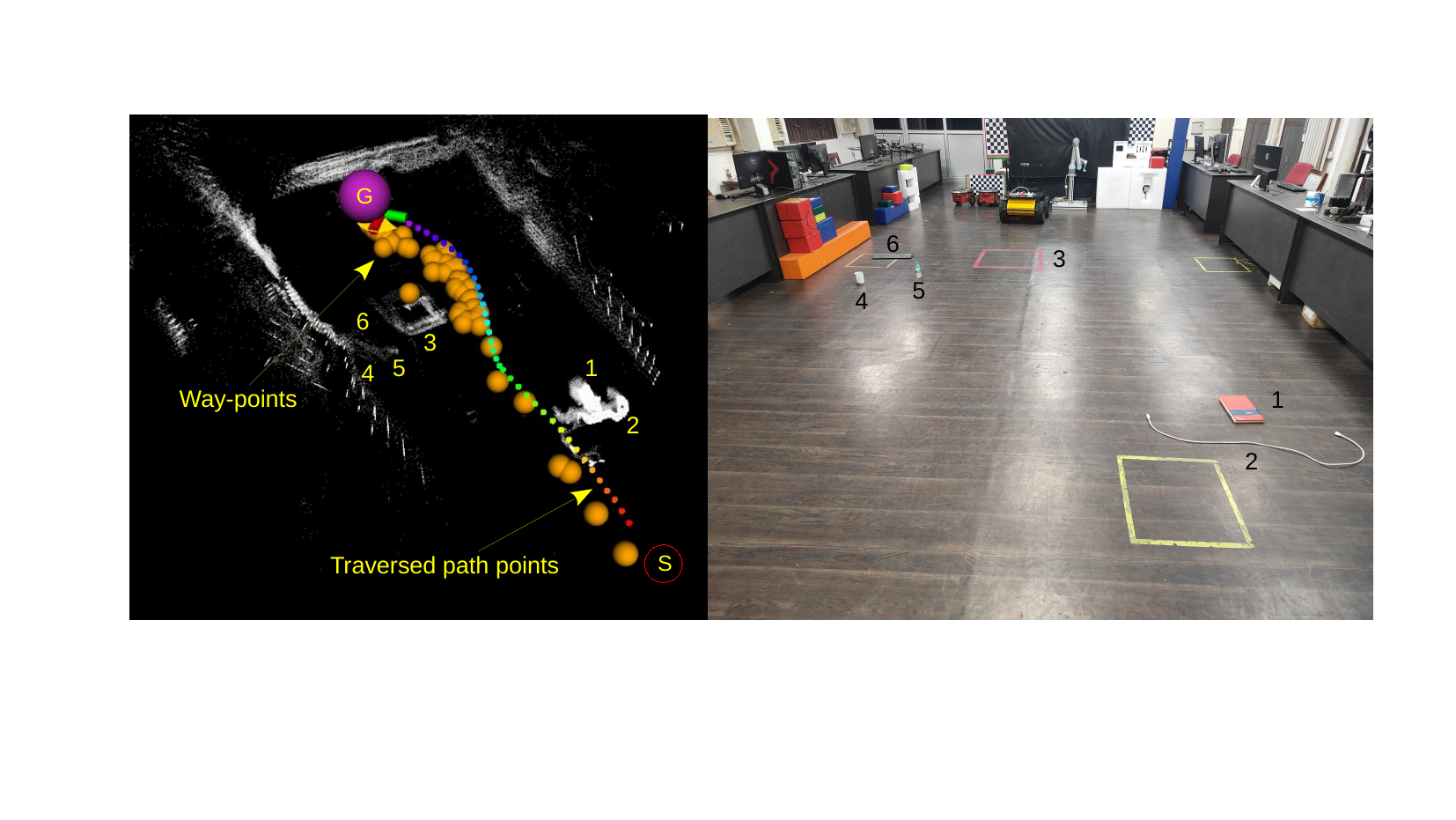}
	\caption{Real-World Navigation Performance: The figure presents our algorithm's trajectory (Left: traversed path and point cloud) within a complex environment (Right: RGB image) avoiding all the geometric and sementic hazards. The successful path demonstrates robust avoidance of geometric obstacles and six "user beware" items (1. book, 2. electric cable, 3. red-marked box, 4. cup, 5. water bottle, 6. keyboard).}
	\label{rw3}
\end{figure}

\begin{figure}[!h]
	\centering
	\includegraphics[width=1.0\columnwidth, trim={7.0cm 4.0cm 8.0cm 4.0cm}, clip]{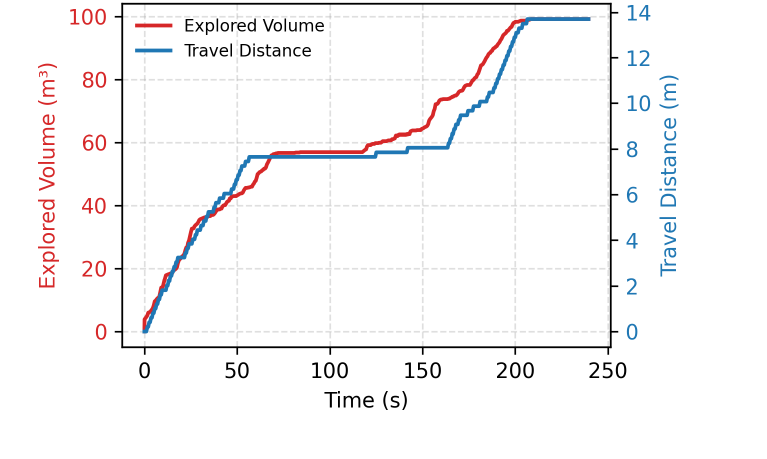}
	\caption{The graphs illustrate the algorithm’s efficiency in navigation and exploration within a complex environment. The proposed method effectively avoids two categories of hazards—physical obstacles and predefined no-go zones (user-beware regions)—as depicted in Fig.~\ref{rw1}~(Right Image).}
	\label{rw4}
\end{figure}
\subsection{Real-World Experiment}
To validate the reliability and efficiency of our proposed method, even under limited computational resources, we conducted experiments using the Clearpath Husky A200 robot (dimensions: 1.0 m × 0.7 m). The platform is equipped with an NVIDIA Jetson AGX computer running our algorithm and relies solely on a single Intel RealSense D455 camera to perceive its surroundings. This setup required the robot to explore cautiously within the constraints of the camera’s narrow field of view (FOV—Depth: 87°×58°, Color: 90°×65°).

Fig.~\ref{rw1} establishes the baseline validation: despite the robot’s large physical footprint and the highly cluttered navigation environment (also evident in Fig.~\ref{rw3}), our algorithm successfully planned and executed a collision-free trajectory, effectively avoiding both fixed physical obstacles and user-defined semantic hazards from the beware list (1. bottle, 2. cable, 3. box, 4. cable, 5. cup). The quality of the generated paths demonstrates the robustness of our approach, even with the inherent limitations of the RealSense sensor’s depth range and accuracy, as well as odometry derived from the wheel encoder (78,000 ticks/m) and IMU.

Minor segmentation noise was observed under varying lighting conditions, primarily due to the reflective flooring; however, this did not significantly affect performance. Despite these challenges, our algorithm maintained reliable operation in indoor environments, as illustrated in Fig.~\ref{rw1} (right) and Fig.~\ref{rw3} (right). The real-world setups in these figures depict the dense navigation scenes populated with user-defined semantic hazards (e.g., book, electric cable, red-marked box, cup, water bottle, keyboard). The left-side images of Fig.~\ref{rw1} and Fig.~\ref{rw3} show the planned path, generated waypoints, traversed positions, and top-down point cloud representation corresponding to the real-world scenes.

Fig.~\ref{rw2} demonstrates the algorithm’s consistent success rate across different start and goal configurations, confirming its ability to autonomously plan and safely reach target destinations. Finally, Fig.~\ref{rw4} presents the quantitative performance results, reinforcing the high efficiency of the proposed approach. Collectively, these real-world experiments decisively establish that our algorithm can handle complex, noisy sensory inputs, interpret user-defined semantic constraints, and achieve safe, context-aware, real-time navigation on a resource-constrained embedded platform.
\section{Conclusion}
This paper addressed the critical gap between cost-effectiveness and semantic intelligence in service robot navigation. We presented a novel, hybrid framework that overcomes the semantic blindness of traditional systems by tightly integrating a lightweight perception module (ESANet) with a robust online A* planner operating on a dynamic occupancy grid. Using only a single, affordable RGB-D sensor, our approach enables context-aware navigation—understanding user-defined visual constraints like keep-out zones or critical items. The core contribution is this fusion of semantic perception with classical planning, creating a unified map that respects both geometric obstacles and non-geometric constraints. Through extensive simulation and real-world experiments on embedded hardware, we demonstrated that our framework achieves robust, real-time performance in unknown and partially known environments. It proves the feasibility of deploying truly intelligent, context-aware navigation on affordable service robots. This work offers a practical and scalable alternative to LiDAR-based systems, highlighting the value of semantic understanding in the planning loop for enhanced safety and flexibility. Future work could explore more complex semantic reasoning and adaptation to different platforms, furthering the deployment of intelligent service robots in the real world.
\bibliographystyle{ieeetr} 
\bibliography{ref} 
\end{document}